\documentclass[conference]{IEEEtran}
\IEEEoverridecommandlockouts
\usepackage{cite}
\usepackage{amsmath,amssymb,amsfonts}
\usepackage{algorithmic}
\usepackage{graphicx}
\usepackage{textcomp}
\usepackage{xcolor}
\usepackage{longtable}
\usepackage{lineno,hyperref}
\usepackage{xcolor,stfloats}
\usepackage{algorithm}
\usepackage{booktabs} 
\usepackage{verbatim}
\usepackage{booktabs}
\usepackage{times}
\usepackage{tikz,xcolor,hyperref}
\definecolor{lime}{HTML}{A6CE39}
\DeclareRobustCommand{\orcidicon}{%
    \begin{tikzpicture}
    \draw[lime, fill=lime] (0,0) 
    circle [radius=0.16] 
    node[white] {{\fontfamily{qag}\selectfont \tiny ID}};    \draw[white, fill=white] (-0.0625,0.095) 
    circle [radius=0.007];    \end{tikzpicture}
    \hspace{-2mm}}
\foreach \x in {A, ..., Z}{%
    \expandafter\xdef\csname orcid\x\endcsname{\noexpand\href{https://orcid.org/\csname orcidauthor\x\endcsname}{\noexpand\orcidicon}}
    }

\def\BibTeX{{\rm B\kern-.05em{\sc i\kern-.025em b}\kern-.08em
    T\kern-.1667em\lower.7ex\hbox{E}\kern-.125emX}}
\begin{document}

\title{Independent Asymmetric Embedding for Information Diffusion Prediction on Social Networks} 

\author{
\IEEEauthorblockN{1\textsuperscript{st} Wenjin Xie\orcidA{}}
\IEEEauthorblockA{\textit{College of Computer and Information Science}\\
\textit{Southwest University}\\
Chongqing, China \\
xiewenjin@email.swu.edu.cn}
\\
\IEEEauthorblockN{3\textsuperscript{rd} Tao Jia\orcidC{}}
\IEEEauthorblockA{\textit{College of Computer and Information Science}\\
\textit{Southwest University}\\
Chongqing, China \\
tjia@swu.edu.cn}
\and
\IEEEauthorblockN{2\textsuperscript{nd} Xiaomeng Wang\orcidB{}\IEEEauthorrefmark{2}}
\IEEEauthorblockA{\textit{College of Computer and Information Science}\\
\textit{Southwest University}\\
Chongqing, China \\
wxm1706@swu.edu.cn\\
\IEEEauthorrefmark{2}Xiaomeng Wang is the corresponding author.}
}

\maketitle

\begin{abstract}
The prediction for information diffusion on social networks has great practical significance in marketing and public opinion control. It aims to predict the individuals who will potentially repost the message on the social network. One type of method is based on demographics, complex networks and other prior knowledge to establish an interpretable model to simulate and predict the propagation process, while the other type of method is completely data-driven and maps the nodes to a latent space for propagation prediction. Existing latent space design and embedding methods lack consideration for the intervene among users. In this paper, we propose an independent asymmetric embedding method to embed each individual into one latent influence space and multiple latent susceptibility spaces. Based on the similarity between information diffusion and heat diffusion phenomenon, the heat diffusion kernel is exploited in our model and establishes the embedding rules. Furthermore, our method captures the co-occurrence regulation of user combinations in cascades to improve the calculating effectiveness. The results of extensive experiments conducted on real-world datasets verify both the predictive accuracy and cost-effectiveness of our approach.
\end{abstract}


\section{Introduction}
Information diffusion through social networks like Twitter and Facebook has profoundly affected our social, economic, and political environment, due to its convenience and efficiency. The diffusion of information on social networks is a complex and dynamic process, which aroused the great research enthusiasm of researchers. The researches on information diffusion play an important role in many fields such as predicting how popular a piece of information will become \cite{shen2014modeling, feng2020modeling}, finding some nodes in a social network that could maximize the spread of influence \cite{cheng2013staticgreedy, dupuis2020real}, how much a cascade will grow \cite{cheng2014can, krishnan2016seeing} and so on. In this paper, we study the task of information diffusion prediction. The process that people repost messages others post on social media is considered as an information diffusion process on the social network, and the propagation sequence of users that reflects the flow of information by time stamps is defined as a diffusion cascade, such as $A\rightarrow B\rightarrow C\rightarrow D$. Information diffusion prediction is to foresee the future cascades knowing the observed cascades.

Plenty of works have been proposed to tackle this information diffusion prediction problem. The conventional approaches jointly exploit the explicit network structure and the underlying dynamic mechanism, and predict the diffusion process from the perspective of social network science. However, these works are mainly based on some features, such as temporal information \cite{cheng2014can}, diffusion content \cite{tsur2012s} and network nodes’ interactions \cite{goyal2010learning}. Although these methods have shown significant improvements in the diffusion prediction task, the feature engineering in the prediction process requires much manual effort and extensive domain expert knowledge. The representation learning happens to offer a solution to this problem. Its goal is to automatically learn the feature representations from social diffusion data, so that the network embedding can be effectively applied to the downstream tasks without the complicated feature engineering.

In this paper, we propose an independent asymmetric embedding (IAE for short) method to accurately and effectively learn social network embedding for cascade prediction. Our method distinguishes itself from existing embedding methods in two key aspects. First, different from existing methods where individuals are embedded into just one or two latent space(s), our method embeds each cascade with different seed nodes into different latent spaces. The diffusion source node is embedded into a latent influence space while the successors are into the corresponding latent susceptibility space. In this way, our method possesses the flexibility to capture the asymmetric interpersonal influence that is intuitively expected and empirically observed in information diffusion. Meanwhile, the interaction between cascades from different sources is also considered. Second, our method proposes the concept of dominant combination to capture the co-occurrence regulation of users in cascades and reduce the computation complexity at the same time.

\section{Related Work}
Towards the cascade prediction problem, the prediction methods are generally to model the diffusion process making sure the model matched the observed diffusion cascade, and make predictions based on the model. Early models attempt to predict and understand the dynamic of observed propagation such as independent cascade (IC) \cite{kempe2003maximizing, goldenberg2001talk}  and linear threshold (LT) \cite{granovetter1978threshold, granovetter1983threshold}. Some models improved from IC and LT are proposed later to capture some propagation features like time series \cite{rodriguez2011uncovering}, the spreading speed, and the randomness in the spreading sequence order \cite{ranyj}. All these models rely on the underlying diffusion model in the network and call for prior knowledge of network structure. Other models discard the artificial assumption of the information propagation mechanism. They purely use the local structural characteristics of nodes for embedding, and make predictions by neural network method, like \cite{wang2017topological, cao2021information}. 

The other kind of models are fully data-driven and do not require a specific network structure. These works based on network embedding are proposed, which always learn directly from the diffusion cascades and capture a network representation from them. For example, the content diffusion kernel model \cite{CDK} aims to learn the representation of the whole network based on the observed cascades. Each node in the network would be embedded as a vector in a latent space. The model predicts the information diffusion process based on the distance of vectors in the latent space, where the node closer to the source spreads the information earlier. Embedded-IC \cite{bourigault2016representation} is similar and models the diffusion process based on the closeness between the information sender’s vector and receiver’s vector. In the information-based embedding-based diffusion prediction model \cite{gao2017novel}, the diffusion prediction is transformed into a spatial probability learning task in the latent space, by controlling the distance between users in the latent space to preserve the time series information in the diffusion process. Deep collaborative embedding model \cite{zhao2020deep} collaboratively embed the nodes with a deep architecture into a latent space, which can learn nodes' embeddings with the information of diffusion order. Yet, these embedding-based methods pay less attention to the interference when embedding nodes into latent spaces, which may lead that the embedding vectors of unrelated nodes being very close.

\section{Prediction Model}
In this section, we propose the cascade prediction model and introduce the heat diffusion kernel which models the contamination propensity of any node given a particular information source. Thus the goal of our prediction model turns to learn the diffusion kernel and rank nodes by their positions in the latent spaces. 

\subsection{Diffusion Dynamics in Latent Space}
A social network is composed of a set of users $U=\left(u_{1}, \ldots, u_{N}\right)$ and relationships between them. A message $m$ starts from a source user and spreads to contaminates subsequent infected users along with the links of the network. The purpose of diffusion prediction is to predict the cascade $c_{m}=\left(u_{0}^{m}, \ldots, u_{n}^{m}\right)$ describes to whom and when the message spreads among users. $u_{0}^{m}$ here stands for the source user of the cascade $c_{m}$. The former CDK and PAE models both proposed a concept of latent space and used machine learning technologies to resolve the prediction problem. CDK is that users are embedded into a latent vector space where the information diffusion process is modeled using heat diffusion dynamics \cite{CDK}. The influence between users in social networks is usually asymmetric. Different from CDK where users are embedded into a single latent space, PAE embeds each user into two latent spaces: a latent influence space $X$ and a latent susceptibility space $Y$, which can possess the flexibility to capture the asymmetric interpersonal influence \cite{PAE}. 

\begin{figure}[!htb]
	\centering
	\includegraphics[width=7.5cm]{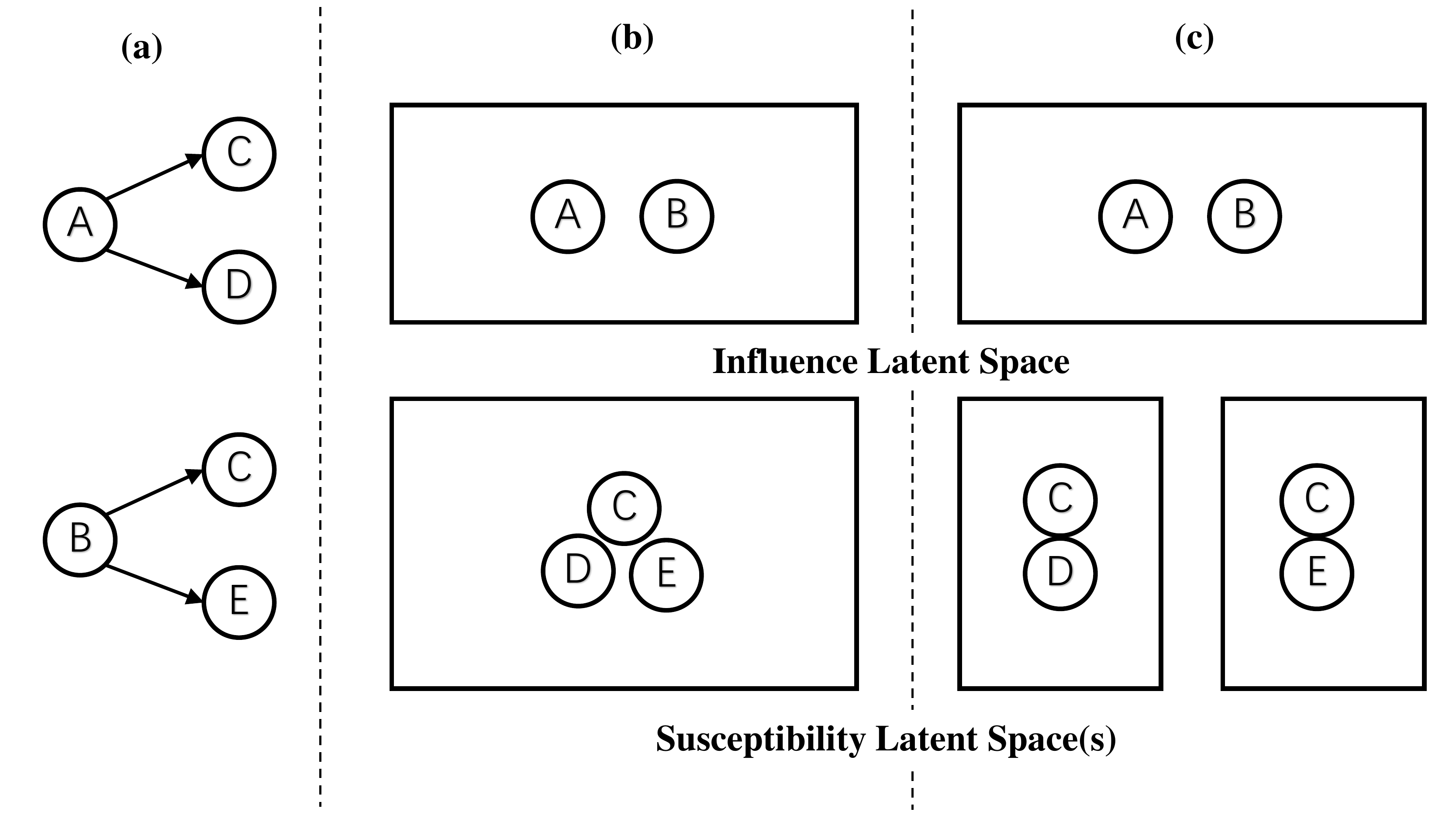}
	\caption{(a) shows a information diffusion case in social network, the arrow means the direction of the message propagates; (b) and (c) illustrate the user embedding circumstances for case (a) in PAE and IAE model separately. Obviously, the unfamiliar users D and E are close in PAE's latent susceptibility space, while they are embedded apart into different spaces in IAE model. 
\label{fig:IAE_strength}}
\end{figure}

In fact, there is still a problem of mutual interference between cascades deriving from different source nodes in the cascade embedding process. This may cause that users who are far apart in the real social network are very close in a latent space yet. For example, in a social network, users C and D are fans of celebrity A, while users C and E are fans of celebrity B, but D and E are not connected. When A sends a message, C and D forward it successively; when B sends a message, C and E forward it successively. In the former models, such a phenomenon will cause the positions of the three users C, D, and E are very close to each other in a latent space, while D and E are not related in the actual network, as shown in Figure \ref{fig:IAE_strength}. And if A publishes two pieces of information, and C, D, and C, and E forward the two pieces of information separately, then there is a high probability that C, D, and E are all fans of A, and there is indeed a relationship between the three. Closely connected, then the prediction results of C, D, and E obtained when predicting the propagation sequence of the message sent by user A will forward the message successively, with a high probability that it is consistent with the actual situation. Therefore, for the cascades with the same source node, embedding cascades into the same latent space is in line with the actual situation. For two cascades from different sources, the users in cascades are supported to be embedded in two latent spaces respectively.

\begin{figure}[!htb]
	\centering
	\includegraphics[width=7.5cm]{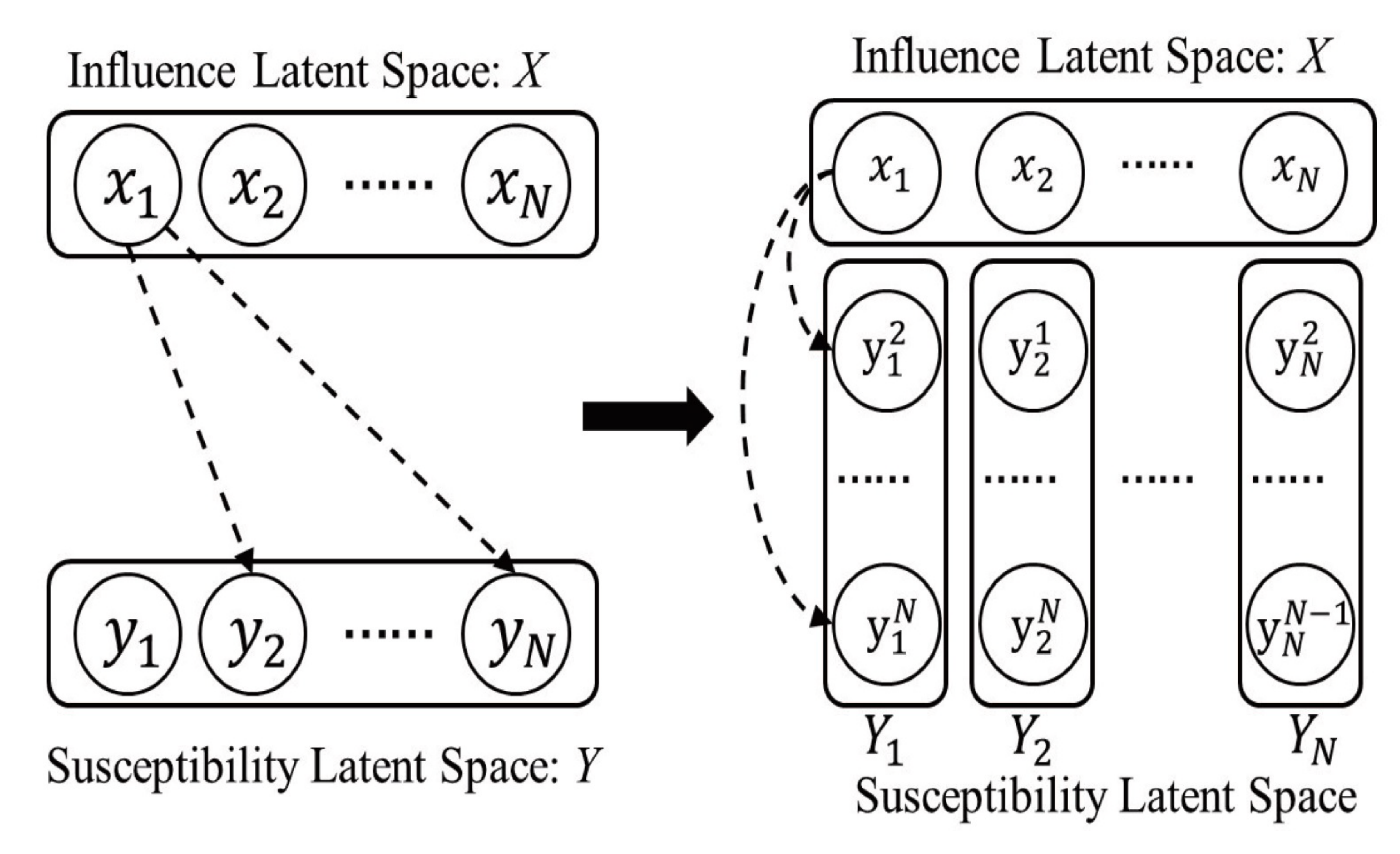}
	\caption{The difference between latent spaces of PAE and IAE models. For a diffusion cascede, its source node will be embedded into an influence space and subsequent infected nodes into one susceptibility space in PAE model, shown as the left. In IAE model, the nodes in the back will be embedded into a special space for their source node.
\label{fig:PAE_IAE}}
\end{figure}

To this end, an independent asymmetric embedding model (IAE) is proposed to avoid interference between source nodes in the latent spaces learning process. For each source node, we build a latent susceptibility space for them respectively and the users in the cascade deriving from this source are embedded into its own latent susceptibility space. IAE embeds users into a latent influence space $X$ and $N$ latent susceptibility spaces $\left(Y_{1}, \ldots,Y_{N}\right)$, so that every user $u_{i}$ as a source has her independent susceptibility space $Y_{u_{i}}$. Figure \ref{fig:PAE_IAE} demonstrates how a cascade will be embedded in these two models. 
	
\subsection{Diffusion Kernel}
The information diffusion process on the social media strictly follows the network structure which means the information can only be disseminated from a user to his neighbor on the social network. In the same way, some works believe the diffusion process of heat is also based on the network structure \cite{kondor2004diffusion, ma2008mining}. Thus the heat diffusion kernel is employed here providing the basis of information diffusion kernel.

The heat diffusion kernel $E\left(t, u_{0}, u_{i}\right)$ computes the heat at location $u_{i}$ at time $t$ knowing that the heat source is $u_{0}$, which defined as

\begin{equation}
E\left(t, u_{0}, u_{i}\right)=(4 \pi t)^{-\frac{n}{2}} e^{-\frac{\left\|x_{u_{0}}-x_{u_{i}}\right\|^{2}}{4 t}}\label{eq:heat_kernel}
\end{equation}

Referring to Equation \ref{eq:heat_kernel}, the new information diffusion kernel can be defined \ref{eq:diffusion_kernel}, in which $x_{u_{0}}$ is location of the source user $u_{0}$ in the latent influence space and $y_{u_{0}}^{u_{i}}$ is user $u_{i}$'s latent susceptibility location in the exclusive space of her source user $u_{0}$. 

\begin{equation}
E\left(t, u_{0}, u_{i}\right)=(4 \pi t)^{-\frac{n}{2}} e^{-\frac{\left\|x_{u_{0}}-y_{u_{0}}^{u_{i}}\right\|^{2}}{4 t}}\label{eq:diffusion_kernel}
\end{equation}

\subsection{Parameters Learning}
According to the diffusion kernel \ref{eq:diffusion_kernel}, the distance from source $u_{0}$ to target $u_{i}$ is the key factor to decide the possibility that $u_{i}$ will be infected by $u_{0}$, a shorter distance $\left\|x_{u_{0}}-y_{u_{i}}\right\|^{2}$ indicate a greater possibility, that provides a strong constraint of our embedding model.

In IAE, the constraint restrict the relationships between the distances from the infected users $(u_i,u_j)$ to the source $u_0$ are defined, which described the situation where user $u_i$ and $u_j$ are both in the cascade $c_m$. And the constraint is defined as

\begin{equation}
\begin{array}{r}
\forall\left(u_{i}, u_{j}\right) \in c_{m} \times c_{m}, t^{m}\left(u_{i}\right)<t^{m}\left(u_{j}\right)\\\Rightarrow\left\|x_{u_{0}}-y_{u_{0}}^{u_{i}}\right\|^{2}<\left\|x_{u_{0}}-y_{u_{0}}^{u_{j}}\right\|^{2} \label{eq:constraint}
\end{array}
\end{equation}
where, $t^{m}\left(u_{i}\right)$ denotes the contamination order number of $u_i$ in $c_m$; $x_{u_0}$ is the influence coordinate of $u_0$ and $y_{u_{0}}^{u_{j}}$ denotes the susceptibility coordinate of $u_0$ in the space $Y_{u_0}$.

Then the diffusion modeling process is to learn the optimal coordinates of the users in the latent spaces from training samples according to the constraint. The corresponding empirical risk of the model is defined as:
\begin{equation}
L=\sum_{c_{m} \in C_{t r a i n}} \triangle\left(E\left( ., u_{0}, .\right), c_{m}\right) \label{eq:risk}
\end{equation}
where $C_{train}$ is the set of training cascades; $\triangle\left(E\left( ., u_{0}, .\right), c_{m}\right)$ is a loss function that measures how much the prediction $E\left( ., u_{0}, .\right)$ given by the diffusion kernel differs from the observed cascade $c_m$. The loss function is defined as follow:
\begin{equation}
\begin{array}{c}
\triangle\left(E\left( ., u_{0}, .\right), c_{m}\right)=\sum\limits_{u_{i} \in c_{m}, u_{j} \in c_{m}, i<j} \max \\
 \left(0, C_{u_{i}, u_{j}}^{m}-\left(\left\|x_{u_{0}}-y_{u_{0}}^{u_{j}}\right\|^{2}-\left\|x_{u_{0}}-y_{u_{0}}^{u_{i}}\right\|^{2}\right)\right)
\end{array} \label{eq:loss}
\end{equation}
where critical penalty margin $C_{u_{i}, u_{j}}^{m}=\log _{\mu}\left(1+\frac{t^{m}\left(u_{j}\right)-t^{m}\left(u_{i}\right)}{1+t^{m}\left(u_{i}\right)}\right)$ is defined in \cite{PAE}. Finally, the problem of learning the optimal coordinates $(X,Y_1,...,Y_N)^*$ becomes:
\begin{equation}
\left(X, Y_{1}, \ldots, Y_{N}\right)^{*}=\underset{\left(X, Y_{1}, \ldots, Y_{N}\right)}{\arg \min } L \label{eq:opt}
\end{equation}
which is to minimize the total empirical loss between the predictions and the observed cascades.

The optimal values for parameter $\{x_{u_0},y_{u_{0}}^{u_{i}},y_{u_{0}}^{u_{j}}\}$ in Equation \ref{eq:loss} can be obtained via minimizing the loss function in Equation \ref{eq:opt}. In this paper, the gradient descent method is used for model parameter estimation and the gradients with respect to $x_{u_0},y_{u_{0}}^{u_{i}}$ and $y_{u_{0}}^{u_{j}}$ are given by
\begin{equation}
\begin{aligned} \frac{\partial L}{\partial x_{u_{0}}} &=2\left(y_{u_{0}}^{u_{j}}-y_{u_{0}}^{u_{i}}\right) \\ \frac{\partial L}{\partial y_{u_{0}}^{u_{i}}} &=2\left(y_{u_{0}}^{u_{i}}-x_{u_{0}}\right) \\ \frac{\partial L}{\partial y_{u_{0}}^{u_{j}}} &=2\left(x_{u_{0}}-y_{u_{0}}^{u_{j}}\right) \end{aligned} \label{eq:gradient}
\end{equation}
The specific learning process is shown as Algorithm \ref{alg:Learning}. 
 
\begin{algorithm}[htb]
\caption{Latent Space Learning Based on GD} 
\label{alg:Learning}
\label{alg:Framwork} 
\begin{algorithmic}[1] 
\REQUIRE ~~\\ 
The cascade set for training, $C$; 
The dimension of latent spaces, $D$;
The maximum number of iterations, $K$; 
The iteration rate, $\eta$;
\ENSURE ~~\\ 
Latent Spaces, $(X,Y_1,...,Y_N)^*$; 
\STATE $k \leftarrow 0$;
\STATE $\left(X, Y_{1}, \ldots, Y_{N}\right)^{k} \leftarrow \text { random }$;
\STATE Extract all dominant combinations $DC=\left\{\left(u_{0}, u_{i}, u_{j}, \overline{c}\right) | t\left(u_{i}\right)<t\left(u_{j}\right)\right\}$ from cascade set $C$, while $\overline{c}$ is critical penalty margin;
\WHILE{$k<K$ and gradients exist} 
\STATE $\left(X, Y_{1}, \ldots, Y_{N}\right)^{k} \leftarrow\left(X, Y_{1}, \ldots, Y_{N}\right)^{k-1}$;
\FOR{$\forall\left(u_{0}, u_{i}, u_{j}, \overline{c}\right) \in C$}
\STATE $\hat{c}=\left\|x_{u_{0}}-y_{u_{0}}^{u_{j}}\right\|^{2}-\left\|x_{u_{0}}-y_{u_{0}}^{u_{i}}\right\|^{2}$;
\IF{$\hat{c}<\overline{c}$} 
\STATE Compute accumulative gradients for $x_{u_{0}}, y_{u_{0}}^{u_{i}}$ and $y_{u_{0}}^{u_{j}}$ using Equation \ref{eq:gradient};
\ENDIF
\ENDFOR
\STATE Update $(X,Y_1,...,Y_N)^k$ with the average value of the accumulative gradients and the iteration rate $\eta$; 
\STATE $\left(X, Y_{1}, \ldots, Y_{N}\right)^{*} \leftarrow\left(X, Y_{1}, \ldots, Y_{N}\right)^{k}$;
\ENDWHILE ~~\\
\RETURN $(X,Y_1,...,Y_N)^*$; 
\end{algorithmic}
\end{algorithm}

\subsection{Prediction Method}
With the estimated coordinates of users in the latent spaces, the model is used to perform cascade prediction on the test cascade set $C_{test}$. The diffusion prediction is treated as a ranking problem. For each cascade $c_m=({u_0},...,{u_k},...,{u_n})$, the users except $u_0$ are ranked in the ascending order by the Euclidean distance $\left\|x_{u_{0}}-y_{u_{0}}^{u_{i}}\right\|^{2}$. The ranked user cascade $c_{m}^{*}=\left(u_{r_{1}}, \ldots, u_{r_{i}}, \ldots, u_{r_{n}}\right)$ implies the diffusion order from $u_0$.

\subsection{Sampling Strategy and Computation Complexity}
The users co-occur in the diffusion cascade is regular, and this co-occurrence association in the network is essential for understanding the internal dynamics of propagation to a great extent \cite{huang2014link, mantymaki2016enterprise, wangxm, battiston2019taking}. In a cascade $(s,...,u,...,v)$ where $s$ is the source user, the user $u$ occurs before $v$, and we use a user combination $(s,u,v)$ to represent this co-occurrence relationship. For an opposite user combination, $(s,u,v)$ and $(s,v,u)$ which may exist simultaneously in the cascade set. If the former combination teems in a cascade set while the latter emerge rarely, it can be inferred that this user order $(s,u,v)$ is almost stable in the social network. Thus it is believed the combination with more occurrences implies the cascading regulation, and we call it the dominant combination. Using the dominant combination in the learning process can reduce the computation complexity while maintaining prediction performance.

Besides, in order to decrease the algorithm computation complexity as well, the same combinations generated from different cascades can be merged and an average critical penalty margin is calculated for the combinations. For example, the combination $(1,3,4)$ appears in the two cascades ${c_1} = (1,5,3,7,4)$ and $c_2 = (1,3,5,7,4)$ and the critical penalty margins of $(1,3,4)$ in the two cascades can be calculated as $C_{3,4}^{1}=0.74$ and $C_{3,4}^{2}=1.32$, then the average critical penalty margin of the combination $(1,3,4)$ is $\overline{c}=1.03$ and the combination should be considered once in an epoch. Here only the dominant combination is sampled for learning.

Thus, the learning complexity of IAE model approaches $O(K \times S \times \overline{C(ni, 2)} \times D)$. K is the number of epochs. S is the number of cascade samples. D is the dimension of the latent spaces. $ni$ is the number of influenced users that is all users in a cascade except the source user and $C(ni,2)$ is the number of dominant user combinations among the influenced users.



\section{Experiments}
In this section, the experiment data and baseline methods are introduced and the results demonstrate the strength of our IAE model.

\subsection{Experiment Data}
The experiments are conducted on the data extracted from real social networks, which are described as follows:

\textbf{Twitter} \cite{twt} is a social media network where people make the information spread by retweet the messages someone else tweets. The dataset extracted from Twitter comprises 137,093 nodes, 3,589,811 edges and 569 cascades (users sharing the same message are treated as a cascade ordered by sharing time).

\textbf{Digg} \cite{digg} is a website where users can submit stories and vote for the stories they like. The dataset contains 279,632 nodes, 2,617,993 edges and 3553 cascades with timestamps.

\textbf{Douban} is a Digg-like social network, which is one of the biggest online social networks in China. The dataset is extracted from the "Top100 users" network, which is composed of the 100 most popular users and their followers contains 13,777 nodes and 567,250 edges. And there are 21,756 cascades in this dataset.

\subsection{Evaluation Metric}
MAP has been used for evaluating cascading prediction \cite{CDK, PAE, wang2017topological, zhao2020deep}. In this paper, we also take MAP as the precision evaluation for prediction performance. The main idea of MAP is to set $n$ cut-off according to the prediction order, calculate the prediction accuracy of each prediction top-k fragments, and finally compute the average accuracy. The calculation framework of MAP is as follows:

\begin{equation}
MAP=\frac{1}{\left|\mathcal{C}_{t}\right|} \sum_{c \in \mathcal{C}_{t}} AP_c
\end{equation}

Where $\mathcal{C}_{t}$ is the cascade set used for testing; $c$ represents a cascade; $AP_c$ represents the prediction accuracy of $c$ and it has different calculation methods according to different understandings. If the top-k precision of $\hat{c}$ is defined as the hit rate of the first $k$ nodes of $\hat{c}$ over the ground truth\cite{zhao2020deep}:

\begin{equation}
AP_c=\frac{1}{|c|}\sum_{{n_k} \in |c|} \frac{|\hat{c}_k \cap c|}{|\hat{c}_k|}
\end{equation}

Where $n_k$ is the node in ground-truth cascade $c$ and $k$ is it's order in the prediction cascade $\hat{c}$. $\hat{c}_k$ is the set of first k objects of $\hat{c}$.

\subsection{Baselines}
Besides the models mentioned above which are CDK and PAE, the other two state-of-the-art network-embedding-based models are also chosen as the baseline methods to compare.



\textbf{Topo-LSTM} \cite{wang2017topological} uses directed acyclic graph as the diffusion topology to explore the diffusion structure of cascades rather than regarding it as merely a sequence of nodes ordered by their infection timestamps. Then it puts dynamic directed acyclic graphs into an LSTM-based model to generate topology-aware embeddings for nodes as outputs. The infection probability at each time step will be computed according to the embeddings.

\textbf{DCE} \cite{zhao2020deep} also investigates capturing the network structure property when embedding. The model uses cascade collaboration to regulate the closeness between any two embeddings of one cascade and use node collaboration to capture the structure similarity between nodes in a cascade when concurrently embedding nodes into latent space. In this way the non-linearity feature of information cascades can be effectively modelled.

\subsection{Experiment Results}
\begin{figure}[!htb]
	\centering
	\includegraphics[width=7.5cm]{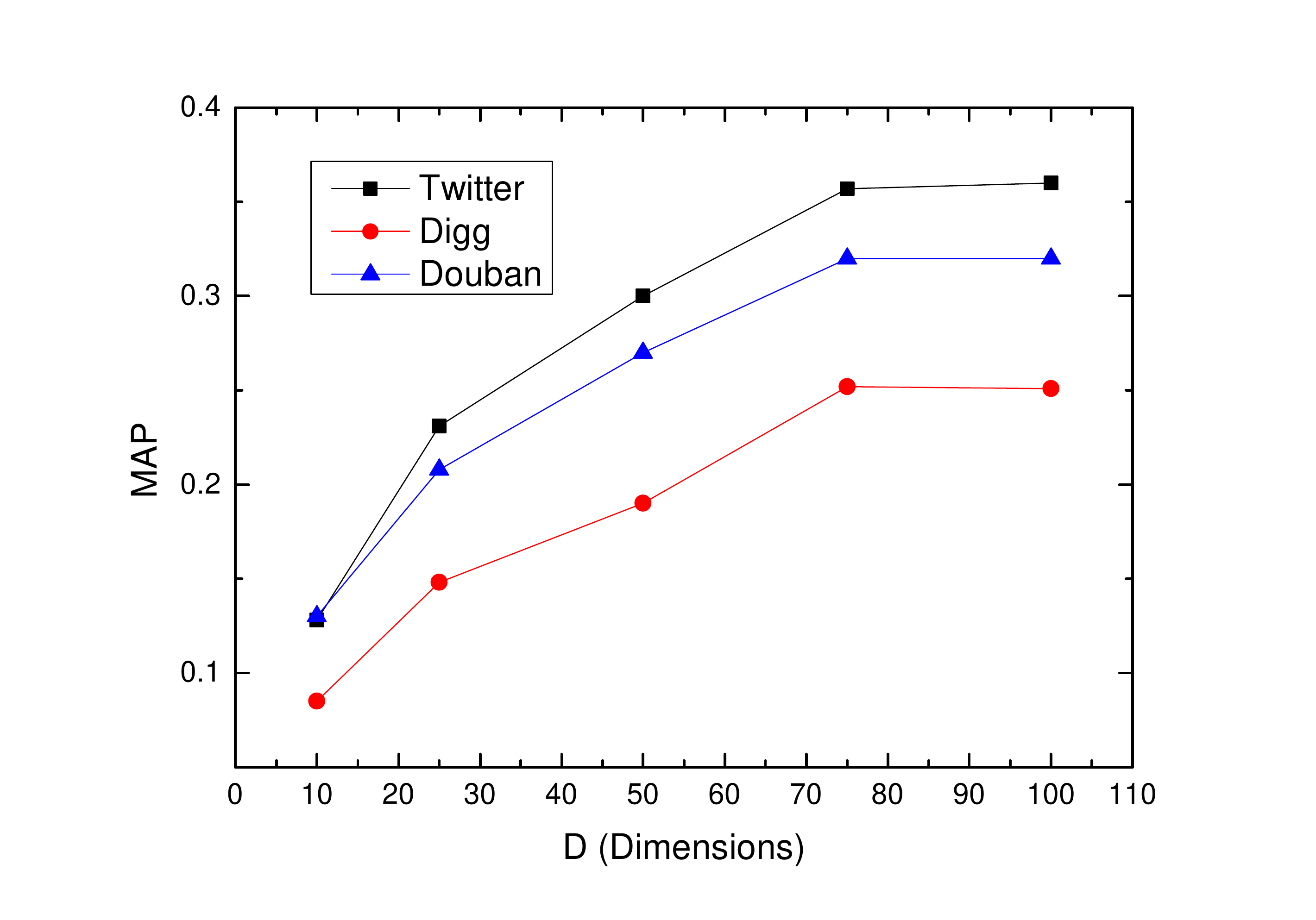}
	\caption{Impact of dimensionality on IAE model.
\label{fig:DIM}}
\end{figure}

We firstly investigate the impacts of hyperparameters. Figure \ref{fig:DIM} demonstrates the prediction accuracy curves of IAE vary with the dimension of latent spaces. All the spaces in IAE share the same dimension. It can be seen that the model reaches the highest MAP when the dimension $D$ in [70,80], thus if no explicit statement, we use $D$ = 75 in the later experiments.  


Then the impact of learning rate $\eta$ is studied. This hyperparameter illustrates the step size of gradient descent during parameter learning. Table \ref{tab:RATE} demonstrates the precision MAP increases when learning rate $\eta$ in [0.0005,0.01], but decreases when $\eta$ continues to grow, which may be attributed to the lack of convergence led by a large learning rate. We use $\eta$ = 0.01 in the later experiments.

\begin{table}[htbp]
  \centering
  \caption{Impact of learning rate $\eta$ on IAE model.}
    \begin{tabular}{c|c|c|c}
    \hline
    $\eta$ & Twitter & Digg  & Douban \\
    \hline
    0.0005 & 0.314 & 0.235 & 0.301 \\
    \hline
    0.002 & 0.327 & 0.259 & 0.312 \\
    \hline
    0.01  & \textbf{0.347} & \textbf{0.282} & \textbf{0.315} \\
    \hline
    0.05  & 0.320 & 0.275 & 0.310 \\
    \hline
    0.25  & 0.286 & 0.218 & 0.267 \\
    \hline
    \end{tabular}%
  \label{tab:RATE}%
\end{table}%

\begin{figure}[!htb]
	\centering
	\includegraphics[width=7.5cm]{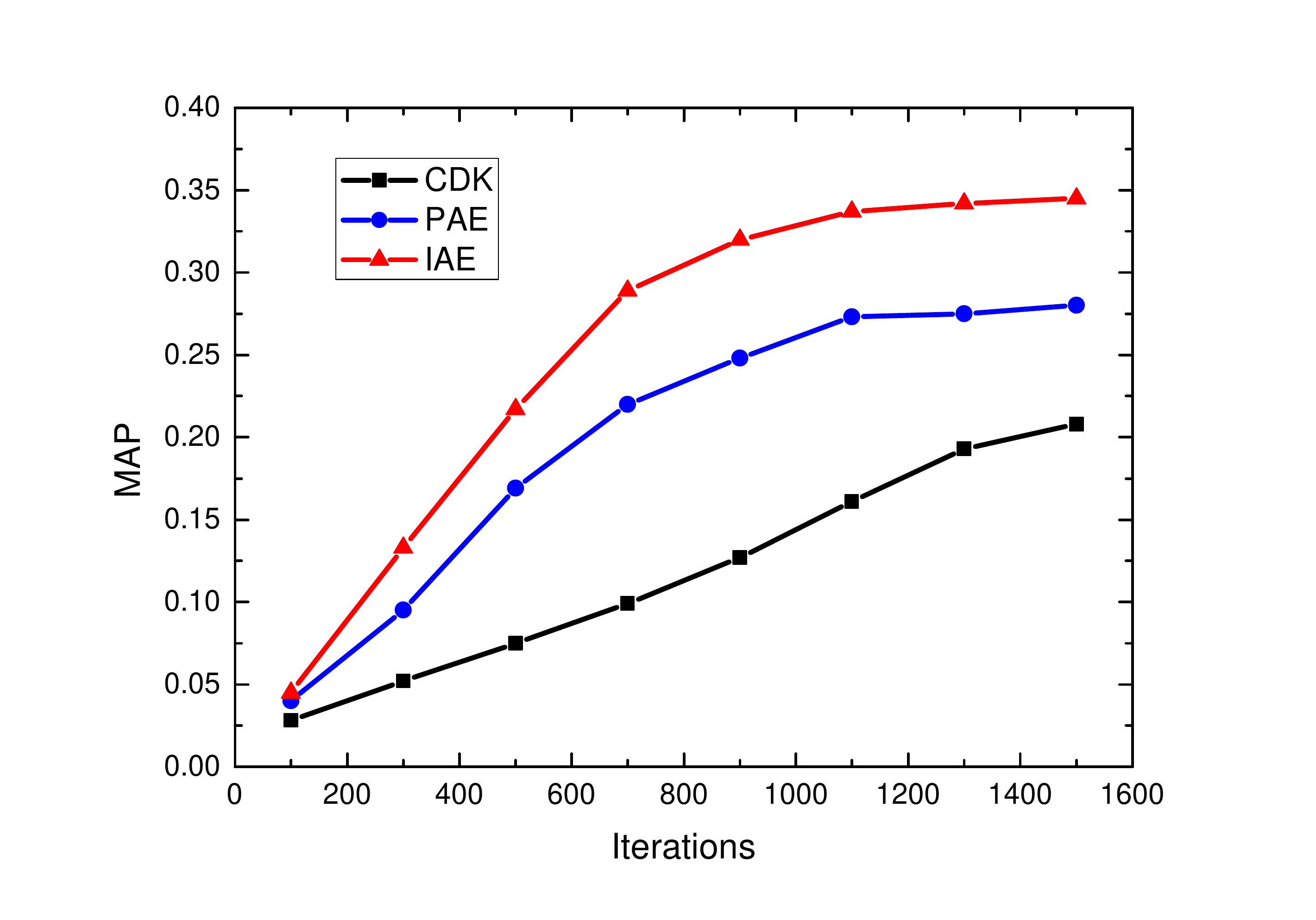}
	\caption{The variation trends of the three models' prediction performances with the increase of iteration steps.
\label{fig:ITERATION}}
\end{figure}

The five models are compared in terms of their accuracy for cascade prediction. And in order to figure out the influence of the exploitation of dominant combination, $IAE_{full}$, which employs all user combinations is also used for comparison. For a fair comparison, we use the best parameter configuration for each method, and the result is shown in Table \ref{tab:MAP-compare}. It can be found that IAE and $IAE_{full}$ outperform others in all three data sets, and $IAE_{full}$ leads a little. But the time consumption of $IAE_{full}$ is near three times of IAE, seeing Table \ref{tab:time}. The models' performance varies with the iteration can also reflect the efficiency. Given that it is meaningful to compare the efficiency of CDK, PAE and IAE, whose design ideas have the continuity relation, only these three models are judged here. As shown in Figure \ref{fig:ITERATION}, the precision increases quickly at the beginning and then tends to be flat, indicating the limit of the model's prediction ability. PAE and IAE quickly achieve the highest MAP comparing to CDK, illustrating the impact of utilizing critical penalty margin. Taking all these terms into consideration, the proposed IAE model gets the best prediction and most cost-effective performance. 

\begin{table}[htbp]
  \centering
  \caption{MAP of the compared models at cascade prediction.}
    \begin{tabular}{c|c|c|c}
    \hline
          & Twitter & Digg  & Douban \\
    \hline
    CDK   & 0.182  & 0.085  & 0.134 \\
    \hline
    PAE   & 0.270  & 0.137  & 0.189 \\
    \hline
    Topo-LSTM & 0.255  & 0.214  & 0.256 \\
    \hline
    DCE   & 0.327  & 0.242  & 0.277 \\
    \hline
    IAE   & 0.358  & \textbf{0.252} & 0.320 \\
    \hline
    $IAE_{full}$ & \textbf{0.374} & 0.251 & \textbf{0.334} \\
    \hline
    \end{tabular}%
  \label{tab:MAP-compare}%
\end{table}%

\begin{table}[htbp]
  \centering
  \caption{Running time (in minute).}
    \begin{tabular}{c|c|c|c}
    \hline
          & Twitter & Digg  & Douban \\
    \hline
    IAE   & 168   & 40    & 45 \\
    \hline
    $IAE_{full}$ & 472   & 98    & 137 \\
    \hline
    \end{tabular}%
  \label{tab:time}%
\end{table}%

\section{Conclusion}
In this paper, the problem of information cascade prediction in online social networks has been investigated using network embedding techniques. Improved from existing methods where the susceptible individuals are embedded into a single latent space, our method embeds each individual into a corresponding latent susceptibility space to avoid the embedding interference between nodes. Furthermore, our method extracts the dominant user combination to capture the cascading regulation and improve the computation efficiency of model prediction accordingly. Thus the proposed independent asymmetric embedding (IAE) model can learn social embedding accurately and effectively for cascade prediction. The results of extensive experiments conducted on real social datasets validate the predictive accuracy and cost-effectiveness of the proposed method.

\section*{Acknowledgment}
This work is supported by the National Natural Science Foundation of China (NSFC) (No.62006198).

\bibliographystyle{IEEEtran}
\bibliography{Independent_Asymmetric_Embedding_for_Information_Diffusion_Prediction_on_Social_Networks}{}

\end{document}